\pdfoutput=1

\documentclass[11pt]{article}
\usepackage[table,xcdraw]{xcolor}
\usepackage{hyperref}
\usepackage[]{EMNLP2022}

\usepackage{times}
\usepackage{latexsym}
\usepackage{graphicx}
\usepackage{hyperref}
\usepackage{tablefootnote}
\usepackage[T1]{fontenc}

\usepackage[utf8]{inputenc}

\usepackage{microtype}

\usepackage{xspace}
\usepackage{inconsolata}
\usepackage{subcaption}
\usepackage{graphicx}

\usepackage{tikz}
\usepackage{calc}
\def\checkmark{\tikz\fill[scale=0.4](0,.35) -- (.25,0) -- (1,.7) -- (.25,.15) -- cycle;} 
\def\scalecheck{\resizebox{\widthof{\checkmark}*\ratio{\widthof{x}}{\widthof{\normalsize x}}}{!}{\checkmark}}

\usepackage{ulem}

\newcommand\crossmark[1][]{%
  \tikz[scale=0.4,#1]{
    \fill(0,0)--(0.1,0) .. controls (0.5,0.4) .. (1,0.7)--(0.9,0.7) ..  controls (0.5,0.5) ..(0,0.1) --cycle;
    \fill(1,0.1)--(0.9,0.1) .. controls (0.5,0.3) .. (0,0.7)--(0.1,0.7) .. controls (0.5,0.4) ..(1,0.2) --cycle;
  }%
}

\def\scalecross{\resizebox{\widthof{\crossmark}*\ratio{\widthof{x}}{\widthof{\normalsize x}}}{!}{\crossmark}}

\title{Too Brittle To Touch: Comparing the Stability of Quantization and Distillation Towards Developing Lightweight Low-Resource MT Models}

 \author{Harshita Diddee\textsuperscript{1} \quad  Sandipan Dandapat\textsuperscript{2} \quad Monojit Choudhury\textsuperscript{1} \\
  \textbf{Tanuja Ganu}\textsuperscript{1} \quad
 \textbf{Kalika Bali}\textsuperscript{1}\\
\textsuperscript{1} Microsoft Research, India \\
\textsuperscript{2} Microsoft R\&D, India \\
{\tt \small \{t-hdiddee,sadandap,monojitc,taganu,kalikab\}@microsoft.com}
}

\usepackage{multicol}
\usepackage{multirow}
\usepackage{booktabs}
\usepackage{amsmath}
\usepackage{amssymb}

\newcommand{\mone}{\textsf{mT5-small}\xspace}
\newcommand{\mtwo}{\textsf{mT5-base}\xspace}
\newcommand{\mthree}{\textsf{mBART}\xspace}

\newcommand{\cpt}{\textsf{CPT}\xspace}
\newcommand{\gmd}{\textsf{GMD}\xspace}
\newcommand{\lam}{\textsf{LA}\xspace}
\newcommand{\kdm}{\textsf{KDD}\xspace}

\newcommand{\kd}{\textsf{KD}\xspace}

\newcommand{\acc}{\textsc{A}\xspace}

\newcommand{\givenmodel}{\textsc{M}\xspace}
\newcommand{\currentmodel}{\textsc{M}\xspace}
\newcommand{\bestmodel}{\textsc{HM}\xspace}

\newcommand{\tfb}{transformer\xspace}
\newcommand{\mt}{mT5\xspace}

\DeclareMathOperator*{\argmax}{\arg\!\max}

\begin{document}
\maketitle
\begin{abstract}
Leveraging shared learning through Massively Multilingual Models, state-of-the-art machine translation (MT) models are often able to adapt to the paucity of data for low-resource languages. However, this performance comes at the cost of significantly bloated models which are not practically deployable. Knowledge Distillation is one popular technique to develop competitive lightweight models: In this work, we first evaluate it's use to compress MT models focusing specifically on languages with extremely limited training data. Through our analysis across 8 languages, we find that the variance in the performance of the distilled models due to their dependence on priors including the amount of synthetic data used for distillation, the student architecture, training hyper-parameters and confidence of the teacher models, makes distillation a brittle compression mechanism. To mitigate this, we explore the use of post-training quantization for the compression of these models. Here, we find that while distillation provides gains across some low-resource languages, quantization provides more consistent performance trends for the entire range of languages, especially the lowest-resource languages in our target set.
\end{abstract}

\section{Introduction}


While NLP has made giant strides in producing more accurate models, these benefits are often not transferred representatively to end-users who would eventually use a language-technology \cite{ethayarajh-jurafsky-2020-utility, caselli-etal-2021-guiding}. Bloated sizes, cumbersome inference times \cite{tao-etal-2022-compression} and a limited set of languages that these models serve are a few reasons for this. More specifically, their usage is hindered by access bottlenecks such as (a) \textbf{Infrastructural Obstacles}: A large percentage of end-users do not have sustained access to internet or high-compute devices to enjoy a stable access to cloud-inferencing of current NLP models \cite{ranathunga2022some, 10.1145/3530190.3534792}, (b) \textbf{Latency Requirements}: Certain NLP services (chat-bots, real-time assistance interfaces, etc.) require very low-inference time which requisite lightweight-models (c) \textbf{Privacy Constraints}: The outflow of sensitive user data which is fed for inferencing to remotely hosted NLP models also has well documented issues \cite{srinath-etal-2021-privacy,huang-chen-2021-improving-privacy, huang-etal-2020-texthide,9286125}. 

Within the research that focuses on evaluating and mitigating these practical constraints, the focus on low-resource language setups has been fairly limited \cite{10.1162/tacl_a_00413}. For instance, while the compression of large language models has received consistent attention through analysis of pruning \cite{behnke-heafield-2020-losing, behnke-etal-2021-efficient}, distillation \cite{bapna2022building,mghabbar2020building,kim-rush-2016-sequence, junczys2018marian} and even quantization \cite{bondarenko2021understanding, 9251854} - much of this work has focused on compressing language models for high-resource languages.

In this paper, we report the results of a comparative analysis of the performance of  distillation and quantization. By focusing on compressing seq2seq multilingual models across a range of languages with data ranging from 7000 to 3M samples - we especially demonstrate the different priors that need to be ascertained for the successful distillation of the model. We are unaware of any previous study that demonstrates the performance of these mechanisms on such low resource languages.

The utility of this work is in commenting on the feasibility of these two compression techniques for rapid development and deployment of MT Models for low resource languages \cite{joshi-etal-2020-state}. More specifically, we believe that distillation's reliance on several priors can be addressed naively through a resource-intensive exercise, where the optimal values of these priors are computed exhaustively. However, in the absence of such a budget, we expect this to be a major impediment in the development of lightweight models for such languages. Since low resource language communities may also be marginalised in other ways, exhaustive investment of data and compute might not be feasible for such communities as well as the language technologists working on these languages \cite{zhang-etal-2022-nlp, 10.1145/3530190.3534792,markl-2022-mind}. 




The main contributions of this work are: 
\begin{enumerate}
\item We distill competitive baseline models for 8 low-resource languages (Bribri, Wixarica, Gondi, Mundari, Assamesse, Odia, Punjabi and Gujarati) and evaluate the sensitivity of the generated models to priors including (a) amount of synthetic Data being used for training (b) The architecture of the student model (c) the training hyper-parameter configuration and (d) the confidence of the teacher models.  

\item We, then, quantize these models to observe if quantization provides a more consistent compression mechanism for these languages. Based on our analysis, we conclude that the suprising stability of naive Post-Training Quantization, especially in the compression of extremely-low resource languages (training data between 5000 and 25000 samples) over distillation. 

\end{enumerate}

We release a combination of lightweight, offline support MT models for these languages along with the scripts for generation and offline inference to further reproducible research in this domain\footnote{\href{https://github.com/microsoft/Lightweight-Low-Resource-NMT}{Codebase and Open-Sourced Models}}.

\begin{table*}[!ht]
    \centering
    \small
    \begin{tabular}{rcrccccccc}
        \toprule
        \multirow{2}{*}{Language} & \multirow{2}{*}{Class} & \multirow{2}{*}{Source Language} & \multicolumn{2}{c}{Data Constraints} & \multicolumn{2}{c}{Model Constraints} \\
        & & & Monolingual Data & Parallel Data & Shared Script &  Included in Pretraining  \\
         \midrule
         Bribri & 0 & Spanish & \scalecross & \scalecheck & \scalecross & \scalecross  \\
         Wixarica & 0  & Spanish & \scalecross & \scalecheck & \scalecross & \scalecross  \\
         Mundari & 0  & Hindi & \scalecross & \scalecheck & \scalecheck & \scalecross  \\
         Gondi & 0 & Hindi & \scalecross & \scalecheck & \scalecheck & \scalecross  \\
         Assammese & 1 & English & \scalecheck & \scalecheck & \scalecheck & \scalecheck \\
         Odia & 1 & English &\scalecheck & \scalecheck & \scalecross & \scalecross \\
         Punjabi & 2 & English &\scalecheck & \scalecheck & \scalecross & \scalecheck \\
         Gujarati & 1 & English &\scalecheck & \scalecheck & \scalecross & \scalecheck \\
         \bottomrule
    \end{tabular}
    \caption{Languages Under Consideration: Note that the except the language's inclusion in the pretraining corpus of our chosen pretrained language models, all factors are independent of our experimental setup. Source language column enlists the source language of the translation pairs}
    \label{tab:languages}
\end{table*}

\section{Approach - Model and Size Adaptations}
In this section, we describe the languages (\ref{languages}), architectures under consideration (\ref{architectures}),  the adaptations that we make for training and fine-tuning these models (\ref{subsec:model-adaptions}) and the adaptations we make to compress their size. 

\label{sec:model-size-adaptations}
\subsection{Languages}
\label{languages}
We perform our analysis on the eight languages shown in Table~\ref{tab:languages}. These languages cover a wide range of availability of monolingual and parallel data, spanning from classes 0 to 3 as defined in ~\citet{joshi-etal-2020-state}. Additionally, they differ in scripts and their inclusion in pretraining corpus which result in interesting modelling adaptions that are needed to be performed for the development of their baselines. In this work, we only study the \textit{High-Resource Language} (\textsc{HRL}) $\rightarrow$ \textit{Low-Resource Language} (\textsc{LRL}) translation direction. The source languages for all our target languages are mentioned in Table~\ref{tab:languages}.


\paragraph{Family of Models}
\label{architectures}
For this work, we leverage two model classes to carry out our analysis: \textbf{I)} seq2seq transformer~\cite{vaswani2017attention}, hereafter referred to as vanilla transformer: With 6 Encoder and Decoder Layers, Vocabulary size - varying between 8k to 32k and 8 attention heads. 
and
\textbf{II)} mT5-small~\cite{xue2021mt5}: With 8 Encoder and Decoder Layers, Vocabulary Size - 250100 and 6 attention heads. 

We train the vanilla transformer from scratch, hereafter referred to as \tfb, to develop a naive baseline for our experiments, and further fine-tune the mT5-small, hereafter referred to as \mt, with certain adaptations for all the languages, as discussed in section~\ref{subsec:model-adaptions}.

For ease of reporting,  we define the highest-performing-model (denoted by \bestmodel) over our family of models as:  

\begin{align}
    \bestmodel = \argmax_{\givenmodel} \acc(\givenmodel) \nonumber
\end{align}

where \currentmodel is a model class with performance $\acc(\currentmodel)$ after training (where \acc is a metric like BLEU \cite{papineni2002bleu} or chrF \cite{popovic-2016-chrf} used to monitor the task-specific performance of the model).


\subsection{Model Adaptations: Language Specific Approaches}
\label{subsec:model-adaptions}
Here we describe the strategies required to adapt these models to different low-resource languages: During fine-tuning, we adapt the pretrained \mt tokenizer to unseen scripts (encountered for Odia) by transliterating it to the closest, highest-resource language included in the pretraining corpus of the pretrained model \cite{khemchandani2021exploiting, ramesh2021comparing, ramesh2022samanantar}.  For our extremely low-resource languages, we used Lexicon-Adaption \cite{wang2022expanding} for the augmentation of target-side monolingual data for languages wherever a bilingual lexicon could be leveraged - Detailed performance with Hindi-Gondi is provided in the Appendix section \ref{gondi-lexicon-adapted}.  However since such methods were not extensible to all the languages in our target language set, we report final experimental results on the models which did not leverage any additional data other than the data mentioned in \ref{data-sources}. Since we analyze the HRL to LRL direction and 4 out of 8 (Bribri, Wixarica, Gondi and Mundari) of our target languages have little to negligible monolingual data - we were also unable to leverage Back-Translation to augment our language-specific parallel corpus \cite{edunov2018understanding}. 

\subsection{Size Adaptation: Knowledge Distillation}
\label{subsec:knowledge-distill}
Knowledge distillation involves training a smaller student network to mimic the token level probabilities of a larger, more accurate teacher model. We distill our models using Hard Distillation \cite{kim-rush-2016-sequence}: we utilize a set of monolingual sentences in the HRL - and forward translate using the \bestmodel to generate synthetic labels that a lighter student model is then trained on.


\subsubsection{Estimation of Optimal Values for Priors}
We define a prior as any attribute of the compression mechanism that needs to be initialized meaningfully and/or optimized for optimal performance - akin to hyperparameters. We use this term specifically so as to put all the dependent variables - such as training data, prediction confidence of the uncompressed models, etc in a single bucket: rather than using a term like hyperparameters that already holds traditional significance in literature. The experimental sweeps for these priors are briefly explained in this section. Note that we focus largely on distillation while estimating for these priors, because quantization provides competitive models even with the default choices established by literature whereas with distillation - the estimation of these priors is critical to achieve a competitive compressed model variant in most cases.   

\paragraph{Prior 1: Optimal Student Architecture}
Following prior work like \citet{bapna2022building}, we experimented with 3 candidate architectures, two of which used deep encoders and shallower decoders. We sweeped across 3 candidate architectures - all variants of a seq2seq transformers with (a) 8 Encoder + 6 Decoder Layers (b) 6 Encoder + 4 Decoder Layers and (c) 6 Encoder + 3 Decoder Layers. We chose the architecture that gave the best BLEU performance after 30 epochs. Sweeps for the architecture were done across each of the following languages - Gondi, Assamesse and Odia as they covered a wide range of training data.  

\paragraph{Prior 2: Optimal Training Hyperparameters}

We sweeped across a set of hyper-parameter sets for Bribri, Gondi, Assamesse and Gujarati to identify the optimal set for the distilled student models. Our goal here was to specifically study the transferability of a hyperparameter set which performed competitively for one or more languages, to all the languages in our target set. 

\paragraph{Prior 3: Amount of Training Data for the Student}
We sweeped across 3 candidate sizes of our synthetic dataset: 100K, 250K and 500K pseudo-labels. Since this decision could also be greatly dependent on the quality of the labels generated per language - we ran this sweep for Bribri, Gondi, Odia and Gujarati, as the quality of the labels generated by the teachers for these languages would be expected to demonstrate significant variation. 

\paragraph{Prior 4: Optimal Teacher Architecture}
To do a preliminary quantification of the effect of the choice of a teacher architecture and the quantity of data that a teacher is trained for on the compressibility of the model - we decided to evaluate the confidence of our teacher models on the predictions they generated. For this, we sampled 100 instances from each of our testsets and monitored the logit distribution of our teacher models.  Specifically, we calculated the average of the softmax entropy of the token-level softmax distributions for a sequence. Taking inspiration from the unsupervised estimation of quality of machine translation outputs \cite{fomicheva2020unsupervised} through similar methods, we hypothesised that the lower the entropy of our model, the more confident it would be in its predictions for a given sample. The intuition here was that if a model is confident about its prediction, its logit distribution would be highly-skewed, and not resemble a uniform distribution (which would indicate its indecisiveness in being able to predict the right token - and therefore, the right sequence). Eventually, this could be used to gauge the quality of the pseudo labels that are student were being trained on. 

\subsection{Size Adaptation: Quantization}
\label{subsec:quantization}
Quantization is a common way to reduce the computational time and memory consumption of neural networks  \cite{wu2020integer}. Here, a lower-bit representation of weights and activation functions is used to achieve a lower memory footprint. In this work, we perform post-training quantization, where after training the base model with full precision of floating point 32 bits (fp-32), we convert the weights and activations of the model to 8 bit integers (int-8). Note that during inference, we still preserve the precision of the input and output encoder-decoder distributions as fp-32. In theory, this brings down the memory consumption of the model by nearly 4x times, though we see an effective reduction of about 3x in practice. More details on the memory-reductions achieved are specified in the Appendix \ref{size-comparison}



    

\section{Experimental Setup}

\subsection{Data}
\textbf{(a) Bribri and Wixarica:} We use the training data 7K and 8K sentences, respectively from \citet{feldman-coto-solano-2020-neural}  and evaluate on test data from \citet{mager-etal-2021-findings}.
\textbf{(b) Gondi}: We use 26k sentences from the data opensourced by CGNET Swara \cite{gondidata} and split it into training and test sets.\footnote{\label{train-test-deduplication}To avoid any test-set leaks, we deduplicate the data by removing tuples ($S^{i}$, $T^{i}$) where $S^{i}$ is the $i^{th}$ sentence in the source language and $T^{i}$ is $i^{th}$the sentence in the target language, between the train and the test set.}
\textbf{(c) Mundari: } We use a dataset of 10K sentences provided by Indian Institute of Technology, Kharagpur\footnote{Data to be released soon;}, and split it into training and test sets.\footnotemark[1]
\textbf{(d) Assamesse, Odia, Punjabi and Gujarati}: We use the training data from \citet{ramesh2022samanantar} (with 0.14M, 1M, 2.4M and 3M sentences, respectively) and evaluate on test data from FLORES200 \citet{goyal2022flores} for Assamese and WAT2021 \citet{wat-2021-asian} for the remaining languages.
Additional details about datasets (sizes and splits) are mentioned in the Appendix \ref{data-sources}.
 


\subsection{Training Setup}


\textbf{Hyperparameters:} We use the \tfb and \mt as our model classes as described previously in Section~\ref{sec:model-size-adaptations}. The hyperparameters for our \tfb model was optimized for fine-tuning of Odia, trained on ~1M sentence pairs. For fine-tuning, we use the Adafactor optimizer \cite{shazeer2018adafactor}, with a linearly decaying learning rate of 1e-3. Since training with smaller batches is known to be more effective for extremely low-resource language training \cite{atrio2022small}, we tuned the training batch size for every language - varying from 32 to 256 (with gradient accumulation as 2) though we did not see very significant variation in the performance on the basis of this tuning.  For our stopping criteria: we fine-tuned all models for 60 epochs (which concluded with considerably overfit models) and then selected models by we picking the checkpoint which had the best validation performance on BLEU (with only the 13a tokenizer which mimics the mteval-v13a script from Moses) \cite{post2018call}. 

We use the sentencepiece tokenizer to build tokenizers for training the baselines for each of the languages \cite{kudo2018sentencepiece}. We use the per-token cross-entropy loss for fine-tuning all our models. Following \citet{xu-etal-2021-vocabulary}, we opt for a relatively smaller vocabulary size with the intent of learning more meaningful subword representations for our extremely low-resource languages. Specifically, we use a vocabulary size of 8K for Gondi, Mundari, Bribri and Wixarica, compared to 32K used for Assamesse, Odia Punjabi and Gujarati. 

\textbf{Experimental Setup for Distillation}
 For Mundari and Gondi we utilize 500K Hindi sentences sampled from the Samanantar corpus \cite{ramesh2022samanantar}; We use the corresponding English corpus to sample English sentences for generating the pseudo labels for Assamesse, Odia, Punjabi and Gujarati. For Bribri and Wixarica - We use Spanish data made available by the Tatoeba Challenge \cite{tiedemann-2020-tatoeba}. We use the per-token cross-entropy loss for training our distilled models.

\textbf{Evaluation Metrics: }  We use BLEU (sacrebleu with spm pre-tokenization (version 2.2.0)) \cite{post2018call} for all our evaluations \cite{goyal-etal-2020-efficient}. In addition to this, we also report chrF2 \cite{popovic-2016-chrf} for all our experiments for a more comprehensive comparison between the models. 


\section{Results}
In section \ref{baseline-model-analysis}, we present the performances of our base models in Table~\ref{tab:highest_performance}. In the following section \ref{compressed-model-analysis}, we report the performances of the distilled \bestmodel in Table~\ref{tab:post_adaptation_performance}. Using these empirical results we focus on answering the following questions (a) To what degree can scaling the student training data improve the performance of the student model? (\ref{subsec:data-sensitivity}) (b) How sensitive is distillation to the choice of the architecture of the student model? (\ref{subsec:student-sensitivity}) (c) How can we choose an optimal teacher that is most suitable for compression? (\ref{subsec:teacher-sensitivity}) (d) To what degree does the hyperparameter set suitable for distilling a model for one language transfer to another language? (\ref{subsec:hyperparam-sensitivity})

While answering these questions, we also analyze in parallel the performance of the quantized variants of these models implicitly indicating the reduced sensitivity of these variants from most of the previously discussed priors in spite of their competitive performances.  

\begingroup
\setlength{\tabcolsep}{4pt} 
\renewcommand{\arraystretch}{1} 
\begin{table}[!h]
    \centering
    \small
    \begin{tabular}{lccccc}
    \toprule
    \multirow{2}{*}{Language} & \multirow{2}{*}{Data}  & \multicolumn{2}{c}{Vanilla \tfb} & \multicolumn{2}{c}{\mt}  \\
    \cmidrule(lr){3-4} \cmidrule(lr){5-6}
    & & spBLEU & chrF2 & spBLEU & chrF2  \\ 
    \midrule
Bribri    & 7K    & 1.7  & 11.6 & \textbf{6.4}  & 19.3   \\   
Wixarica  & 8K    & 2.2  & 14.1 & \textbf{6.2}  & 28.0   \\
Mundari   & 10k   & 0.1  &  5.6 & \textbf{15.9} & 33.7   \\
Gondi     & 26K   & 1.2  & 7.9  & \textbf{14.3} & 32.5   \\ 
Assamesse & 140K  & 0.8  & 12.4 & \textbf{10.7} & 30.4   \\
Odia      & 1M    & 23.7 & 43.6 & \textbf{27.4} & 47.6   \\
Punjabi   & 2.4M  & \textbf{38.4} & 50.6 & 34.8 & 44.1   \\
Gujarati  & 3.05M & \textbf{35.9} & 53.4 & 35.7 & 49.8   \\

    \bottomrule
    \end{tabular}
\caption{Performance of our base models (\tfb and \mt) without quantization or distillation. Best performing models out of the two architectures are marked in bold.}
\label{tab:highest_performance}
\end{table}
\endgroup

\subsection{Analyzing the Baseline Models}
\label{baseline-model-analysis}
As expected, the transformer models for target languages start competing (and outperforming) once an adequate amount of data is available for training the vanilla transformers. In addition to the obvious gain for being only optimized for target languages, the performance gains of these baselines can also be attributed to the language-specific tokenizer that they utilize, in contrast to the pretrained \mt tokenizer that might be sub-optimal for language-specific generation. For our low-resource languages though, the advantage of transfer learning is clearly evident: all languages achieve a minimum and maximum performance improvement of 4 and ~16 BLEU points. Gondi and Mundari, despite having relatively low-amount of data, perform well - though we expect an overestimation of their performance due to the homogenity between the train and the test set. Additionally though, both languages share scripts with a dominant language script i.e., Devanagari and hence, can be expected to gain because of that. 

\subsection{Analyzing the Compressed Models}
\label{compressed-model-analysis}
In Table \ref{tab:post_adaptation_performance}, we briefly present the performances of our distilled and quantized models. As evident, especially for the lowest-resource models, both distillation and quantization give competitive performance in addition to providing a significant size reduction. Note that Table \ref{tab:post_adaptation_performance} does not report the performance of the quantization of the vanilla transformer models for Odia, Gujarati and Punjabi even though they had competed or outperformed the \mt variants. This is because they suffered a significant drop in performance - Odia dropped in performance to 8.4 BLEU/30.5 chrF2 in contrast to its \bestmodel scores of 23.7 BLEU/ 43.6 chrF2 respectively. Gujarati and Punjabi also dropped to 16 BLEU/31.2 chrF2 and 19.1/36.0 , respectively. 
To explain this we note what distinguishes these two architectures: (a) \mt is deeper than \tfb having 2 extra layers on the encoder's side than the vanilla transformer and (b) leverages multilingual pretraining. These attributes become useful in interpreting \mt robustness to compression. In agreement with prior work like \citet{li2020train}, deeper models can be expected to be more immune to compression. In fact, these models can be expected to be regularized by a certain degree through quantization, and we posit that we might be adopting a sub-optimal fine-tuning hyperparameter set for the initial fine-tuning of these models, consequently generating potentially overfit models and this gets mitigated to some extent upon quantization. Taking into consideration the lack of prior work on fine-tuning large LMs on such extremely low-resource languages and the infeasibility of running intricate hyperparameter sweeps per language with such large models, this can also be expected to degrade the quality of the labels generated for training the distilled models - ultimately affecting the performance that the distilled models achieve.


\begingroup
\setlength{\tabcolsep}{3pt} 
\renewcommand{\arraystretch}{1} 
\begin{table}[h]
    \centering
    \small
    \begin{tabular}{lccccc}
    \toprule
     \multirow{2}{*}{Language} & \multicolumn{1}{c}{\bestmodel} & \multicolumn{2}{c}{Distilled \bestmodel} & \multicolumn{2}{c}{Quantized \bestmodel}  \\
     \cmidrule(lr){2-2} \cmidrule(lr){3-4} \cmidrule(lr){5-6}
     & spBLEU & spBLEU & chrF2 & spBLEU  & chrF2  \\
    \midrule
    Bribri    & 6.4   & 6.8  & 13.2 & \textbf{7.4}  & 19.4  \\ 
    Wixarica  & 6.2   & 4.1  & 17.3 & \textbf{7.2}  & 26.8  \\
    Mundari   & 15.9  & \textbf{18.2} & 32.7 & 15.7 & 29.3  \\
    Gondi     & 14.3  & \textbf{14.2} & 32.8 & 13.8 & 31.1  \\
    Assamesse & 10.7  & \textbf{9.6}  & 27.4 & 6.2  & 25.7  \\
    Odia      & 27.4  & 20.2   & 40.7 & \textbf{21.0} & 41.3  \\
    Punjabi   & 38.4  & \textbf{32.8} & 46.6 & 27.0 & 48.0  \\
    Gujarati  & 35.9  & \textbf{29.8} & 48.6 & 28.4 & 51.4  \\
    \bottomrule
    \end{tabular}
    \caption{Performance of the \bestmodel for all languages after applying Distillation and Quantization. Best performing models out of both of the size adaptations are marked in bold. }
    \label{tab:post_adaptation_performance}
\end{table}
\endgroup

In the following sections we focus on presenting our analysis of distillation's sensitivity to certain priors. In each section, we also discuss an analysis of the same priors' effect on quantization. Note that since the mT5 outperformed the vanilla transformer variants for all languages up till Odia - we distilled and quantized them for these languages. Also note that the \bestmodel for these languages is hence, \mt.  Additionally, for Odia, Gujarati and Punjabi, we quantized both the mT5 and the vanilla transformer variants of the models.


\subsection{Sensitivity to Priors: Data} The quality, quantity and the domain of data that the teacher or uncompressed variant of the model is trained on, appears to impact both the mechanisms of compression: For distillation the gold training data as well as the monolingual data utilized for generating student labels is of relevance, and for quantization only the gold data that the teacher is fine-tuned for, is of relevance.

\paragraph{Quantity of Training Data}
Interestingly, quantization displayed consistent performance variations across the entire range of our low-resource language sets (all languages up till Odia), giving marginally close scores to the \bestmodel so at least within the data sparse languages we did not see any direct variation in the performance according to the amount of training data used. Both mechanisms show nearly equal degradation in performance for the HRL.  
\label{subsec:data-sensitivity}

\paragraph{Quality of Training Data}
The quality of the data that the teacher is trained on affects the model's immunity to compression. This is best demonstrated by the post-compression performances of Gondi and Mundari in Table \ref{tab:post_adaptation_performance}: In Gondi - the train set has nearly 26K sentences, which by the virtue of being collected via crowd-sourcing may be expected to be noisy. Mundari's training data, though also crowd-sourced, claims to have been validated manually after its collection by the providers to generate the final corpus of about 10K sentences. The observed difference where Gondi suffers a slight performance degradation post-compression and Mundari experiences a significant performance gain, may be attributed to the difference in the quality of their training data. Note that both languages are being translated from the same source language, share the same script and are being tested on a correlated test set - so the quality and quantity of training data are expected to be major contributors to the variations in their performance.\footnote{The two languages do belong to two different language families - Gondi belonging to the Dravidian language family which has a higher representation in the pretraining corpus for mT5, and Mundari being Austro-Asiatic}

\begin{figure}[!h]
    \centering
    
    \begin{subfigure}{0.65\columnwidth}
        \includegraphics[width=\columnwidth]{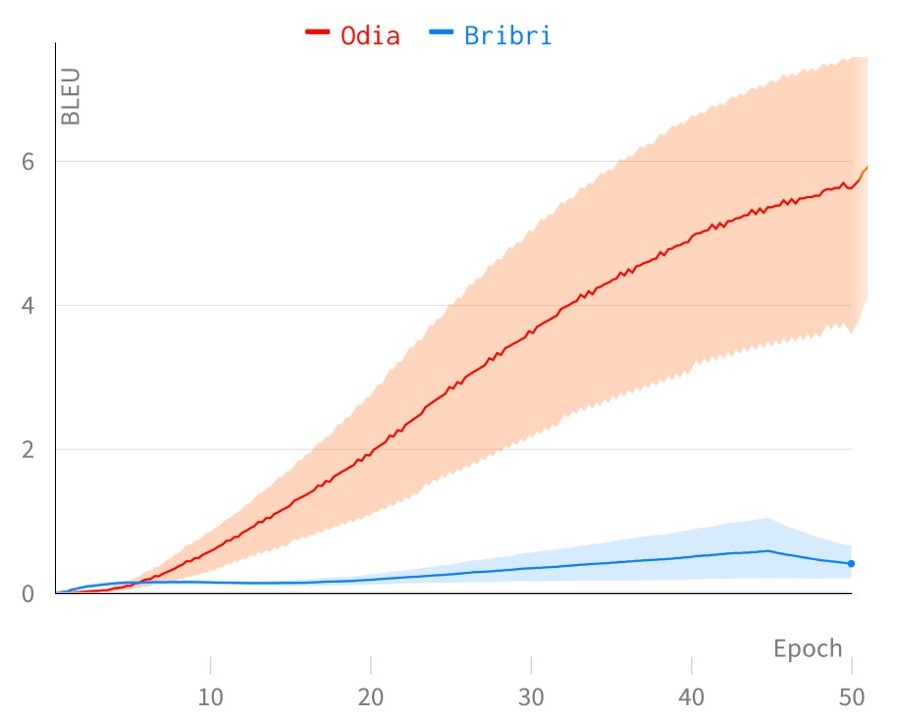}
        \caption{Variation in the efficacy of pseudo-labels between Bribri and Odia}
        \label{fig:first}
    \end{subfigure}
    
    \hfill
    
    \begin{subfigure}{0.65\columnwidth}
        \includegraphics[width=\columnwidth]{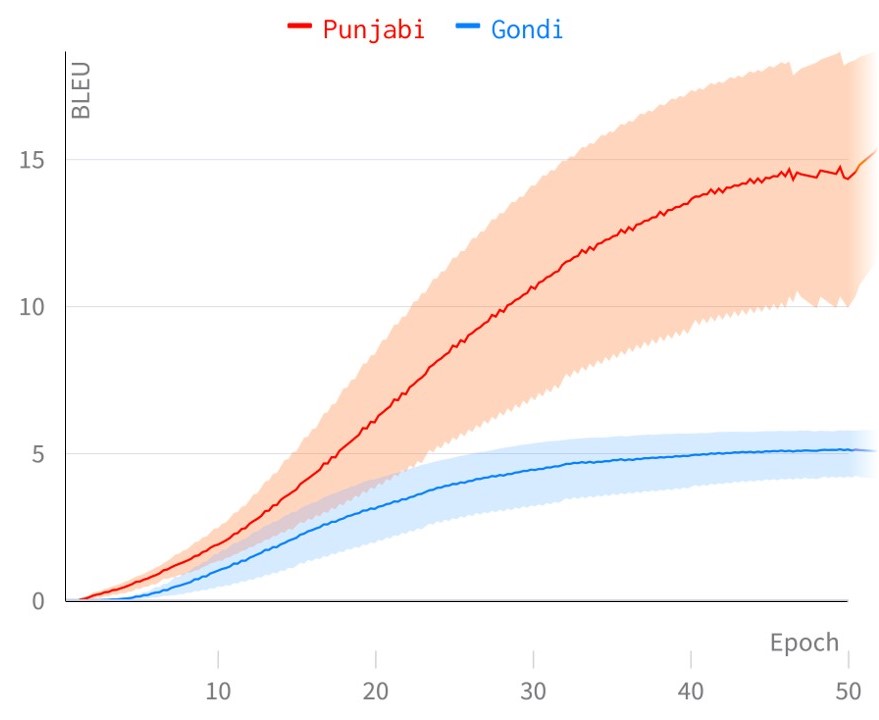}
        \caption{Variation in the efficacy of pseudo-labels between Punjabi and Gondi}
        \label{fig:second}
    \end{subfigure}
\caption{Min/Max range curves of the performance of the models trained on scaled data: The shaded range is considerably lower for the lowest-resource languages indicating reduced efficacy of scaling student data.}
\label{fig:sensitivity_data}
\end{figure}

\paragraph{Quantity of Pseudo-Labels used for Student's Training}

 Results of our analysis of scaling student data between 100K to 500K are presented in Figure \ref{fig:sensitivity_data}. More data seemed to help for the entire spectrum of languages - though it is evident that the gain in the performance diminished in proportion to the amount of added data as we approached the lowest-resource languages in our set. The gain in performance upon the addition to 250K samples to a HRL like Odia or Punjabi is significantly more pronounced than the gain in performance for Bribri or Gondi - where there is a very marginal improvement in the performance upon the addition of 250K samples. This could be indicative of the diminishing efficacy of the increasingly noisy data that was generated by the lowest-resource teachers. We explore this notion in more depth in Section~\ref{subsec:teacher-sensitivity}.

\paragraph{Domain of Data} While we do not perform any targeted experiments to evaluate the domain dependence of the two compression mechanisms - we posit that the distilled models' significantly better performance than its quantized variant in Assamesse could be attributed to the distilled model's exposure to the diverse-domain data during the student's training. Note that the testset used in Assamesse, FLORES 200 \cite{goyal2022flores}, is claimed to be of a very diverse-domain origin. Given this, the process of training a student on monolingual data of a potentially more diverse origin to that of the native training set - would explain the gain that the language demonstrates on a domain-agnostic testset. Prior work like \citet{mghabbar2020building} already shows distillation's efficacy in enabling students to adapt to out-of-domain data that the teacher may not have ever been exposed to. Quantization on the other hand, has no opportunity for exposure to any out-of-domain data - so its adaptation and performance across a domain-agnostic testset can be expected to only degrade.

 
\subsection{Sensitivity to Priors: Student Architecture}
\label{subsec:student-sensitivity}

\begin{figure}[h]
\centering
\includegraphics[width=\columnwidth]{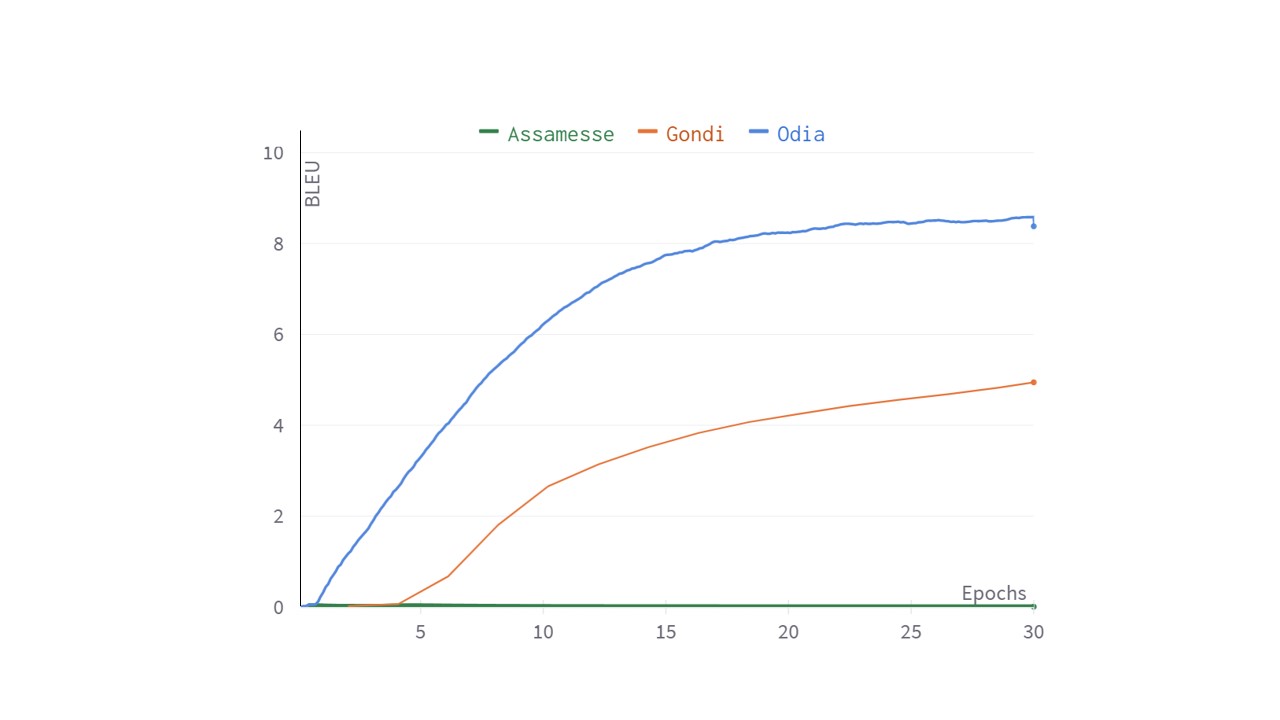}
\caption{Variation in BLEU due to difference in the choice of a student architecture: An optimal architecture choice for Odia and Gondi gives adversely sub-optimal performance for Assamesse}
\label{student_arch}
\end{figure}

We find that distilled student models could be adversely sub-optimal for a given language, despite being sub-optimal or even an optimal choice for a large subset of languages. To demonstrate this in Figure \ref{student_arch}, we show the performance of two distilled models on an identical hyperparameter set and student architecture. While the chosen student architecture gives competitive performances for Gondi and Odia, Assamesse performs significantly worse for this candidate architecture. We did attempt retraining the model with a different seed to negate the possibility of a randomly poor initialization though this did not improve the convergence. While we did not notice such a drastic performance variation across any other candidate set, this instance did indicate brittleness to the student-architecture for a given language.  After these sweeps, we fixed a transformer-based encoder with 6 layers and a transformer-based decoder with 4 layers as the distilled model for our further experiments. 


\subsection{Sensitivity to Priors: Confidence of the Teacher Model}
\label{subsec:teacher-sensitivity}

\begin{figure}[h]
    \centering
    \includegraphics[width=0.8\columnwidth]{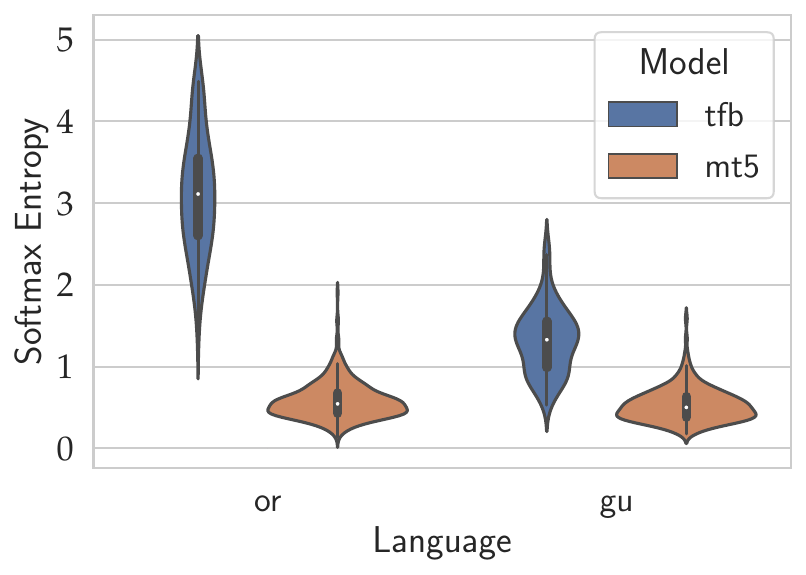}
    \caption{Entropy distributions of \mt and \tfb: lower-entropy indicates high-confidence and consequently suggest higher-quality of translations.}
    \label{fig:mt5_vs_tfb}
\end{figure}

Estimating the confidence of our teacher models displayed manifold benefits: Within Distillation, it helped us get an indirect estimate of the quality of the training data that the student model was trained on. Within Quantization, it was useful in analyzing why the mT5-variants were more robust to quantization. Note that since the testsets for all the languages are of varying difficulty - doing a language-wise comparison on the basis of such metrics was non-trivial since the confidence predictions could also vary in accordance with the complexity of the testsets being evaluated upon. Hence, we majorly focused on analyzing languages which were either evaluated on the same test set (Gujarati, Punjabi, Odia with WAT21 testset \cite{wat-2021-asian}) or the different architectures for each of our languages which could be evaluated for the same testset.

Figure \ref{fig:mt5_vs_tfb} demonstrates the difference in the entropy of the softmax distributions of the mT5 and transformer teacher variants. Note that this is for Gujarati and Odia, our highest resource language, for which both architectures perform quite competitively and the vanilla transformer even outperforms the mT5.

As is evident, the mT5 variant has much lower entropy scores, with lower dispersion indicating high-confidence in the predictions it produces for each of the samples. Note that the inference pipeline for both architectures is identical - Greedy Search with no sampling so we don't expect any difference in the decoding mechanism to affect the quality or distribution of representations that we are monitoring. This is a very interesting observation, as both models appear to perform comparably according to our automatic metric evaluations - yet differ quite significantly in the stability with which they generate these predictions.

    \begin{figure}[!h]
    \centering
    \includegraphics[width=0.8\columnwidth]{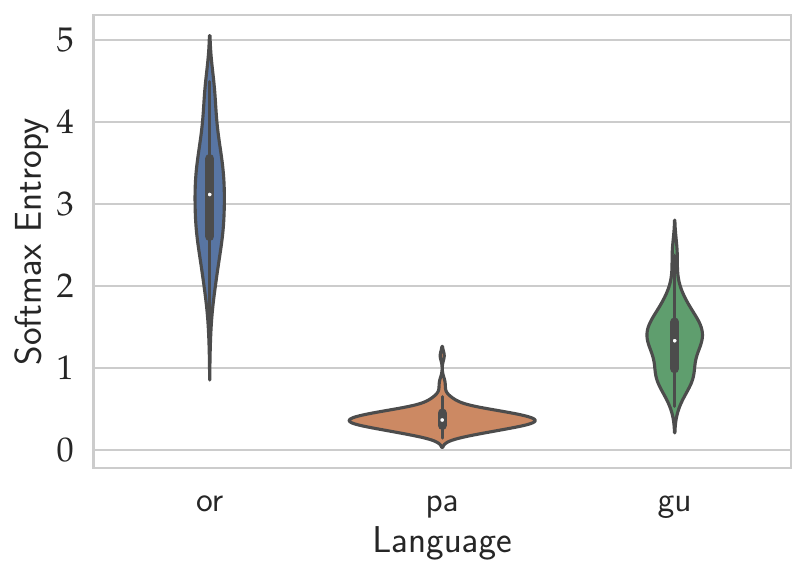}
    \caption{Entropy distributions for \tfb across different languages: Models become increasingly more confident about their predictions with an increase in training data}
    \label{fig:or_pa_gu}
    \end{figure}

Next, we attempt to establish if training with more data makes a model more confident in its prediction. Figure \ref{fig:or_pa_gu} demonstrates the entropy scores for Odia, Punjabi and Gujarati. Each of these have data increasing in the order of 1M, 2.4M and 3M respectively. Here we observe that indeed, models trained with more data achieved consistently lower entropy scores.


\subsection{Sensitivity to Training Hyperparameters}
\label{subsec:hyperparam-sensitivity}
In this section we present results of evaluating if an adequate hyperparameter set for a given language may be suitable for generating an optimal variant for another distilled language. Here too, we demonstrate using a subset of our hyperparameter sweep that there can be a marked degradation in the suitability of an averagely optimal hyperparameter set (that might be close to optimal to multiple languages with similar attributes) to an unseen language;

    \begin{figure}[ht]
    \centering
    \includegraphics[width=0.8\columnwidth]{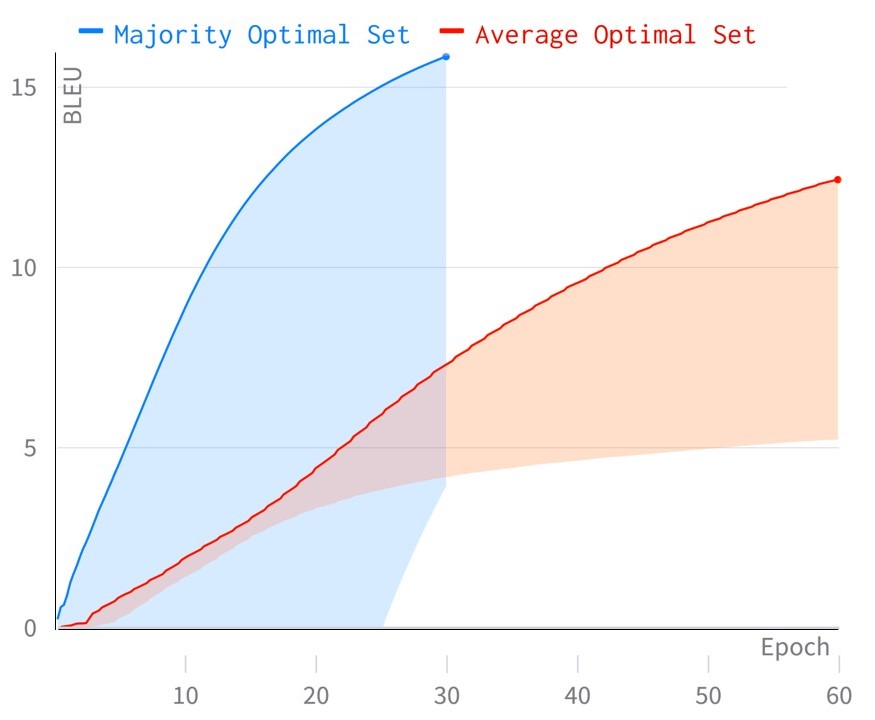}
    \caption{Min/Max range of performances of Gujarati, Bribri and Assamesse across a hyperparameter set that is optimal for these languages but adversely sub-optimal set for Gondi}
    \label{fig:sensitivity_of_hparams}
    \end{figure}
In Figure \ref{fig:sensitivity_of_hparams}, when tuned for the hyperparameter set that is optimal for a majority of languages in our set, Gondi does not even converge as a result of which the lower-bound of a teacher's performance for that hyperparameter set is 0. Note that this hyperparameter set transferability does not seem to show any specific data oriented trends as well. For instance, the same hyperparameter set that was optimal for Gujarati, our highest resource language with 3M data points, is only slightly sub-optimal for Bribri, our lowest resource language with 7000 data points, and Assamese, our mid-resourced language with 135K sentences. Also note that we were able to get acceptable performance for Gondi with almost an identical hyperparameter setup with a larger batch size (quadrupled to the one in this setup) indicating that a per-language sweep would be an ideal and acceptable solution even though this would imply that distilling models would mandate a significant hyperparameter tuning for achieving optimal performance. A detailed list of what hyperparameters we sweeped through can be found in the Appendix Table \ref{table:hyperparameter_candidates}.

\section{Takeaways}
We encapsulate the learning from our analysis as the following takeaways:  
\begin{enumerate}
  
    \item \textbf{\textit{Data Dependence of the Method of Compression:}} Training teacher models with lesser quantity, higher quality data is expected to improve a model's robustness against both quantization and distillation. The post-quantization performance suffers equally for models trained with varying degrees of data. This is not the case with distillation, where increasing the amount of training data for student distilled models starts providing diminishing returns as the amount of training data for the teacher reduces.
    
  
    \item \textbf{\textit{Cost of Compression:}} Distillation is quite sensitive to its training hyperparameters and the student's architecture. This choice doesn't necessarily follow any data-oriented trends as well i.e., languages having similar amount of data may perform very differently on similar hyperparameter and student architecture sets. Hence, Distillation mandates a significant hyperparameter tuning cost that Quantization does not incur.   
    
    \item \textbf{\textit{Stability of Compression}:} Hard Distillation and Post-Training Quantization are both promising methods of quickly compressing massively multilingual models for machine translation for extremely low-resource languages. Post-Training Quantization should be preferred when the uncompressed variants is pretrained and/or deep, expected degree of compression is upto 4x the original model's size and the cost of compression is to be minimum. Distillation, on the other hand, should be preferred when domain-expansion, language-specific tokenization and more than 4x degree of compression needs to be achieved at the cost of a tuning for optimal architecture and training setup selection. 
     

\end{enumerate}

\section{Related Work}

Owing to the known benefits of compressing language models due to their
lower-memory footprint, improved inference speed and even improved performance in some cases, compression techniques have been explored widely in NLP. 

\paragraph{Quantization}

While the work on quantizing encoder-models is replete \cite{9463531,bondarenko2021understanding,kim2021bert,9251854} the focus on quantizing decoder-only models \cite{tao2022compression}, and specifically seq2seq models has been relatively much lower. Recent work like, EdgeFormer, \cite{ge2022edgeformer}, LLM.int8() \cite{dettmers2022llm} have recently demonstrated the generation of seq2seq quantized models which provide a high-compression ratios and competitive performances though this work has also been done with much higher resource languages. 

\paragraph{Distillation}
Work within distillation is replete, even for the multilingual-type of models that we focus on. Work like \citet{kaliamoorthi2021distilling,jiao2021lightmbert,yang2022cross} represent the major body of work in multi-lingual distillation - that is also centered across the encoder-only space. Relatively lesser work has been done in the space of mutli-lingual distillation \cite{soltan-etal-2021-limitations,mukherjee2021xtremedistiltransformers} of seq2seq models and even though work like \citet{zhang2020improving,he2019language} extends this analysis to relatively low-resource languages, they rely on the use of monolingual data for the target language, a luxury that we cannot afford for half of the  languages in our language set. 

Note that since both processes are orthogonal, their conjunctive use has also been explored - \citet{tao-etal-2022-compression} for instance, get competitive results by applying token level contrastive distillation and module-wise dynamic scaling while quantizing generative models. Note that we made the conscious decision of excluding pruning from our analysis because while it is known to demonstrate very effective parameter reduction, it is generally not as aggressive in it's memory footprint reduction as much as quantization and distillation \cite{behnke-heafield-2020-losing, mohammadshahi2022compressed}. As we'll discuss further in section \ref{section:discussion}, size-reduction was an implicit focus of this work that is one of the most fundamental bottlenecks of community deployment \ref{size-comparison}.

\section{Discussion}
\label{section:discussion}

While this work explicitly focuses on only the performance comparison between distillation and post-training quantization, it's efficacy can also be viewed in demonstrating the development of lightweight, machine translation models for extremely low-resource languages. This is a very critical outcome as Performance-oriented Machine translation (MT) models for low-resource languages are often not suited for the immediate consumption of the community.  The access bottleneck introduced by these bloated models, can especially affect those communities which haven't traditionally enjoyed access to a digital ecosystem, often widening the gap between those who can and cannot access these tools. Towards this direction, the exploration of compression strategies for these models - especially when tied to end-user centric NLP services such as translation is imperative. In this work, the size of all models being evaluated after compression was less than 400MB - the quantized models are at least 3x lighter the size of the native \bestmodel and the distilled models give even more impressive gains of upto 8x smaller than their uncompressed counterparts. This size reduction, coupled with the increased speed of inference associated with this reduction in most cases can enable a suite of accessible translation models for these languages\footnote{A more detailed description of the sizes of these models and the associated inference patterns is provided in the Appendix \ref{size-comparison}}.  This establishes a very promising potential in achieving deployment-constraint aware models: For instance, in areas where users do not enjoy a sustained access to the internet - these light-weight models may be adapted to operate on edge in an offline fashion. 

\section{Conclusion and Future Work}

In this work we established that hard-distillation is sensitive to several priors which makes it a brittle mechanism of compression, especially for languages with extremely low-resources. In relative comparison, post-training quantizaton provides a competitive, stable and cost-effective compression mechanism that works effectively for extremely low-resource languages as well. Moving forward, we wish to explore the effect of using additional data (augmented or natively available) on the compressed variants of these models and extend distillation's analysis to utilizing logit distributions of the teacher (soft-distillation). Having observed the poor confidence measures of the transformer - and it's relatively random distributions we expect to get more interpretable evidence towards the suitability of these models for soft distillation through such an analysis. 

\section*{Acknowledgements}
We sincerely thank the reviewers for their detailed feedback on the work which greatly helped us improve the quality of this work. Additionally, we thank Indian Institute of Technology, Kharagpur for giving us access to the Mundari data.  Finally, we also thank Anurag Shukla for helping formalize the quantization pipeline and Kabir Ahuja, Sumanth Doddapaneni and Naman Jain for all the helpful discussions.  

\bibliography{anthology,custom}
\bibliographystyle{acl_natbib}

\appendix
\section{Appendix}

\subsection{Details of Data Sources}
\label{data-sources}
For all the languages in Table ~\ref{tab:languages} we now describe the training and evaluation corpora used. Note that for languages like Assamesse, Odia, Punjabi, etc. we could have accessed a monolingual corpus to supplement our training as well but since we wouldn't have been able to leverage data at a similar scale and quality for the entire language set, we abstained from using methods that leveraged monolingual corpora in these languages.  
\paragraph{Bribri} Training data from \citet{feldman-coto-solano-2020-neural} containing about 7K parallel sentences. Test data from \citet{mager-etal-2021-findings} with 1003  sentences. 
\paragraph{Wixarica} Training data from \citet{feldman-coto-solano-2020-neural} containing about 8k parallel sentences. Test data from \citet{mager-etal-2021-findings} with 1K sentences. 
\paragraph{Mundari} We requested Indian Institute of Kharagpur for Data on Mundari. This corpus contained 10K parallel sentences. We partition train and test sets from this and generate a test set of 980 sentences \footnote{\label{footnote:test-leak-dedup} To avoid any test-set leaks, we deduplicate the data by removing tuples ($S^{i}$, $T^{i}$) where $S^{i}$ is the $i^{th}$ sentence in the source language and $T^{i}$ is $i^{th}$the sentence in the target language, between the train and the test set.}
\paragraph{Gondi} Data obtained from \citet{gondidata} containing 26K sentences.
We partition train and test sets from this and generate a test set of 730 sentences\ref{footnote:test-leak-dedup}.
\paragraph{Assamesse} Train data obtained from  \citet{ramesh2022samanantar} containing 0.14 parallel sentences. Test data from \cite{goyal2022flores} containing 1012 sentences 
\paragraph{Odia} Train data obtained from  \citet{ramesh2022samanantar} containing 1M parallel sentences. Test set from WAT2021 \cite{wat-2021-asian} containing 2390 sentences
\paragraph{Punjabi} Train data obtained from  \citet{ramesh2022samanantar} containing 2.42M parallel sentences. Test set from WAT2021 \cite{wat-2021-asian} containing 2390 sentences
\paragraph{Gujarati} Train data obtained from \citet{ramesh2022samanantar} containing 3.05M parallel sentences. Test set from WAT2021 \cite{wat-2021-asian} containing 2390 sentences

\begin{table*}[h]

     \centering
\begin{tabular}{ l c c c c }
\hline
\textbf{Model} &\textbf{Data}  & \textbf{spBLEU} & \textbf{S(\currentmodel)} (in MB)  \\
\hline
Transformer & $26.2k$  & 1.4    & 240      \\

\mone  & $61.9k$  & 12.7 &     1200     \\
\mone  & $26.2k$  & 14.3 &     1200      \\
\mtwo & $26.2k$  & 15.6  & 2100           \\
\mthree & $26.2k$  & 13  &  2280     \\
\mone : \cpt \{\gmd\} & $26.2k^{mono}$ & 14.9  & 1200 \\
\mone : \cpt \{\lam\} & $200k^{mono}$ & \textbf{14.9} & 1200 \\
\mone : \cpt \{\lam\} & $200k^{mono}$ & \textbf{10.8} & 400 \\
\mone : \cpt \{\kdm\} & $143k^{mono}$ & 15.2 & 1200 \\
\mone : \cpt \{\gmd + \lam + \kdm\} & $26.2k + 343k^{mono}$ & 14.7 & 1200 \\
\mone: Quantizing M1 & $26.2k$  & 13.8 & 400 \\
Quantizing CPT Model \{Best \mone\} & $26.2k$  & 10.2  & 400  \\
Transformer + \kd  & $26.2k + 240k$ & 10.1 & 185 \\
\hline 

\end{tabular}
\caption{Gondi: Use of Lexicon Adaptation, Continued Pretraining and Mixed-training with Lexicon Adapted and Forward Translated Monolingual Data.}
\end{table*}

\subsection{Evaluating Continued Pretaining with Synthetically Augmented or Lexicon Adapted Monolingual Data for improving the \bestmodel }
\label{gondi-lexicon-adapted}
 The use of continual pretraining with monolingual data has been shown to be very useful in improving the transfer for low-resource languages. In our cases, our lowest resource languages, i.e, Bribri, Wixarica, Gondi and Mundari, did not have any monolingual data available natively so we explored the augmentation of the same using lexicons \cite{wang2022expanding}. We also generated forward translated data using the \bestmodel that we developed to fuse with the lexicon-adapted data.  
 For continued pretraining we use a fixed learning rate of 0.001. 
Results of our experiments are logged in Table 5.We use the following notations to report our results \textit{\gmd - Gold Monolingual Data, \lam - Lexicon Adapted Monolingual Data, \kdm - Knowledge Distilled Monolingual Data } where \gmd indicates the target-side monolingual data available within the parallel corpus of the language, \kdm indicates the forward-translated data that we generate via our best-performing model for Gondi i.e., mt5-base. We generated 100K labels using mt5-base teacher, and also experimented adding 100K sentences from a weaker teacher, i.e., mt5-small in hopes of leveraging a more diverse class of labels to train the student on. 

We did observe a small gain in performance upon the addition of LA data during pretraining - though the post-quantization performance and the distilled model' significant performance degradation called for a deeper investigation on the effects of continued pretraining for this language. 
 
\subsection{Hyperparameter Trial Configurations}
We ran Hyperparameter sweeps with the configurations specified in Table \ref{table:hyperparameter_candidates}. 

\begingroup
\setlength{\tabcolsep}{3pt} 
\renewcommand{\arraystretch}{1} 
\begin{table}[h]
    \centering
    \begin{tabular}{lc}
    \toprule
    \textbf{Hyperparameter} & \textbf{Candidate Values}  \\
    \midrule
    Train batch size        & 32, 64            \\
    Epochs                  & 10, 30, 60        \\
    Method                  & grid             \\
    Metric                  & BLEU             \\
    Gradient Accumulation   & 2, 4             \\
    Label Smoothing         & 0, 0.1           \\
    Learning Rate           & 5\{e-5,e-5,e-6\} \\
    Warmup Steps            & 500, 1000     \\
    \bottomrule
    \end{tabular}
    \caption{Candidate values of hyperparameters: Sweep for finding the optimal hyperparameter set for Distillation}
    \label{table:hyperparameter_candidates}
\end{table}
\endgroup

\begin{figure}[!h]
    \centering
    
    \begin{subfigure}{0.9\columnwidth}
        \includegraphics[width=\columnwidth]{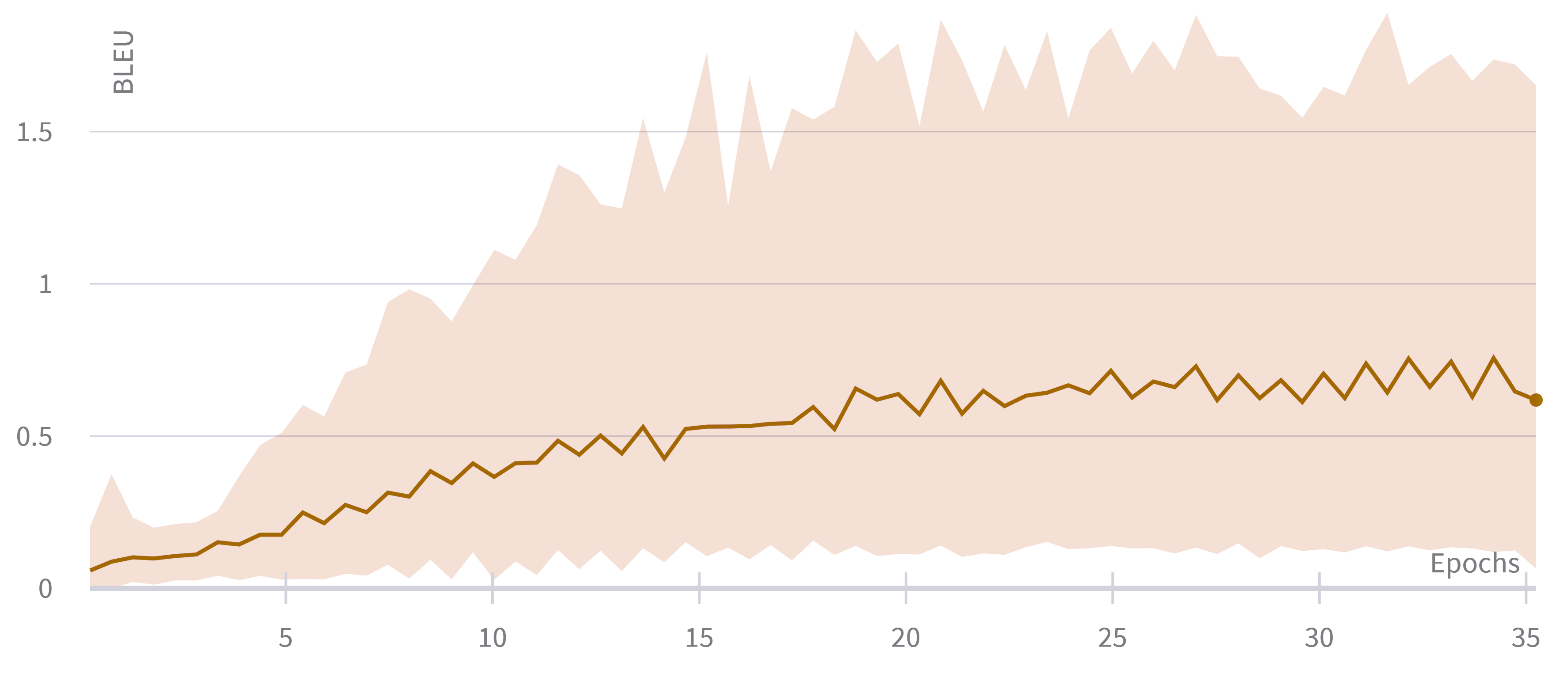}
        \caption{Min/Max Range of Bribri's Sweep}
        \label{fig:sweep1}
    \end{subfigure}
    
    \hfill
    
    \begin{subfigure}{0.9\columnwidth}
        \includegraphics[width=\columnwidth]{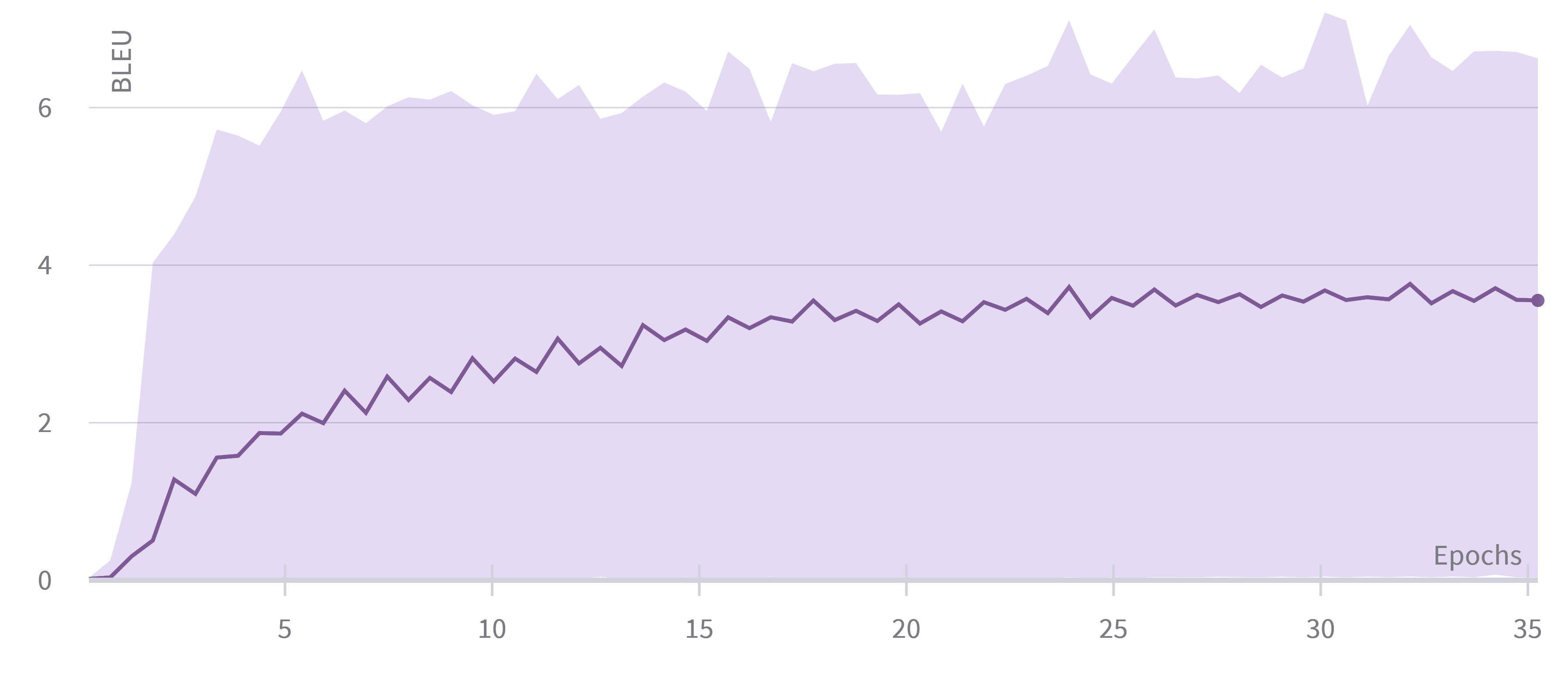}
        \caption{Min/Max Range of Gondi's Sweep}
        \label{fig:sweep2}
    \end{subfigure}
\caption{Variation of performance across languages}
\label{fig:hyp_sensitivity_extended}
\end{figure}

Note that in congruence with the observations of subsection \ref{subsec:hyperparam-sensitivity}, we also provide the min-max range of performance for Gondi and Bribri in Figure \ref{fig:hyp_sensitivity_extended}.

As can be observed, for a set of hyperparameters, at least one of which is optimal for some other language in the set, both languages fail to converge. Similarly, in extension to subsection \ref{subsec:student-sensitivity}, we also checked if for the same hyperparameter set, the variation in student architecture produced significant performance variations.

\begin{figure}[!h]
    \centering
    
    \begin{subfigure}{0.4\textwidth}
        \includegraphics[width=\columnwidth]{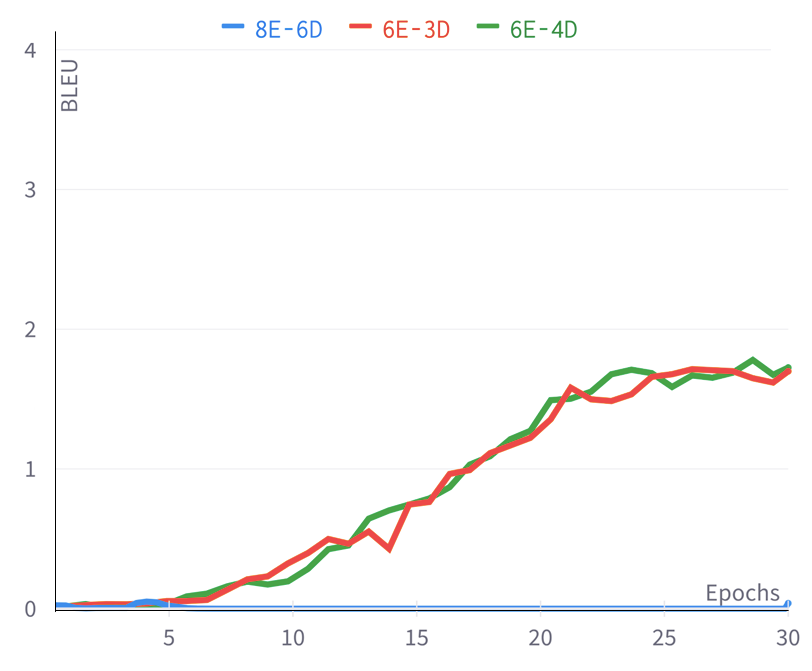}
        \caption{Variation in BLEU with change in student architecture for Assamesse}
        \label{fig:st_as}
    \end{subfigure}
    
    \hfill
    
    \begin{subfigure}{0.4\textwidth}
        \includegraphics[width=\columnwidth]{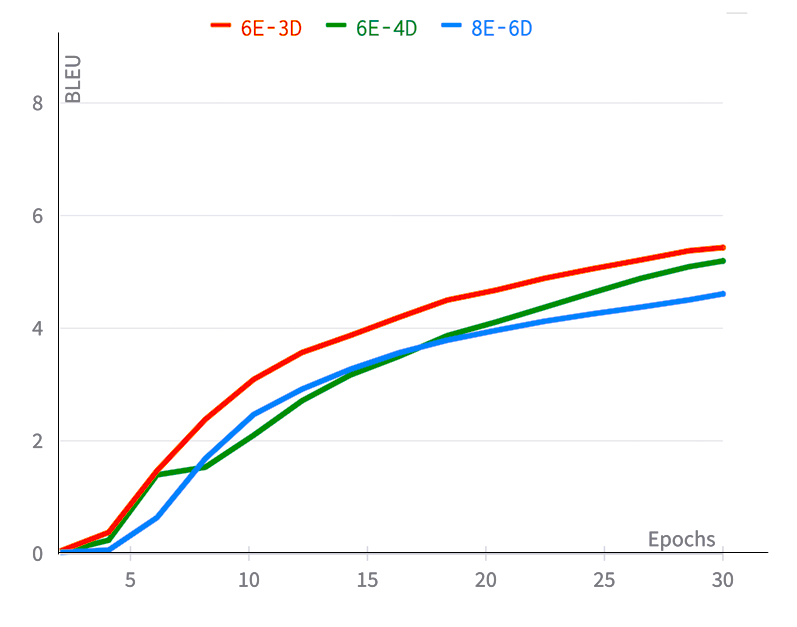}
        \caption{Variation in BLEU with change in student architecture for Gondi}
        \label{fig:st_gondi}
    \end{subfigure}
    
\caption{In the legend E and D refers to Encoders and Decoders respectively}
\label{fig:diff_arch_same_lang}
\end{figure}
    
The results demonstrated in Figure \ref{fig:diff_arch_same_lang} did not show any significant variation except for the case of Gondi, i.e., altering the student architecture - while keeping all other priors the same: adversely affected the performance in that one case. 


\subsection{Comparing Size-Reduction Affinity of Quantization and Distillation}
\label{size-comparison}
 This exploration is extremely useful as the size of a model significantly impacts several factors associated with the consumption of any service, impacting it's adoption by community members through several ways including \textit{(a) Accessibility on Edge:} Since mobile devices are constrained in their RAM and Memory Usage - users with edge devices of low-capabilities are naturally inhibited to is services that drain their device's resources. \textit{Inadequate Connectivity Requirement for Inference, One-time download and Service Updates:} Users may often avoid downloading apps that seem too large, particularly in emerging markets where devices connect to often-spotty 2G and 3G networks or work on pay-by-the-byte plans \footnote{https://developer.android.com/topic/performance/reduce-apk-size}.  \textit{Large Rendering Time:} Finally, a bloated size may often be associated with a larger rendering response period which might hinder the usability experience of a user engaging with the MT service.

\begingroup
\setlength{\tabcolsep}{3pt} 
\renewcommand{\arraystretch}{1} 
\begin{table}[h]
    \centering
    \small
    \begin{tabular}{lcc}
    \toprule
    \textbf{Language} & \textbf{Native S(\bestmodel)} & \textbf{Compressed S(Q,D)}\\
    \midrule
    Bribri        & 1228    & (400, 153)    \\
    Wixarica      & 1228    & (400, 153)          \\
    Gondi         & 1228    & (400, 153)       \\
    Mundari       & 1228    & (400, 153)       \\
    Assamesse     & 1228    & (400, 189)       \\
    Odia          & 1228    & (400, 189)       \\
    Punjabi       & 232     & (75, 189)    \\
    Gujarati      & 232     & (75, 189)     \\
    \bottomrule
    \end{tabular}
    \caption{Sizes of the Uncompressed and Compressed Variants for all languages - Q and D indicate the compressed sizes of the Quantized and the Distilled Models respectively. All sizes are in MB.}
    \label{table:compressed_sizes}
\end{table}
\endgroup

\paragraph{Note on Inference Times}
In theory, compression through both distillation and quantization is expected to be conducive to faster inference for the models: The distilled models are not bounded to use a pretrained embedding and hence can gain in inference by using smaller, target-language specific embeddings. The quantized models can also benefit due to the reduced precision in which the inference operations are carried out, though this optimization is heavily dependent on if the hardware running the model can leverage these operations in their expected precision \cite{bondarenko2021understanding}. Especially in the case of quantization, the scope of this analysis would be quite vast, which is why we also excluded it from our current analysis.


    



\end{document}